\documentclass[conference]{IEEEtran}
\IEEEoverridecommandlockouts
\usepackage{cite}
\usepackage{amsmath,amssymb,amsfonts}
\usepackage{algorithmic}
\usepackage{graphicx}
\usepackage{textcomp}
\usepackage{xcolor}
\usepackage{balance}
\usepackage{textcase}
\usepackage{subfig}
\usepackage[flushleft]{threeparttable}
\usepackage{booktabs}
\usepackage{multirow}
\usepackage{array}
\usepackage{url}
\usepackage{hyperref}
\hypersetup{
    colorlinks=true,
    linkcolor=red,
    filecolor=magenta,      
    urlcolor=blue,
    pdfpagemode=FullScreen,
}

\begin{document}

\title{
Learning Multi-Step Manipulation Tasks from 
\\A Single Human Demonstration

}

\author{
\IEEEauthorblockN{Dingkun Guo}

\IEEEauthorblockA{
\textit{Robotics Institute}, 
\textit{Carnegie Mellon University}, Pittsburgh, PA, USA \\
edu@dkguo.com}
}

\maketitle

\begin{abstract}
Learning from human demonstrations has exhibited remarkable achievements in robot manipulation. However, the challenge remains to develop a robot system that matches human capabilities and data efficiency in learning and generalizability, particularly in complex, unstructured real-world scenarios. We propose a system that processes RGBD videos to translate human actions to robot primitives and identifies task-relevant key poses of objects using Grounded Segment Anything. We then address challenges for robots in replicating human actions, considering the human-robot differences in kinematics and collision geometry. To test the effectiveness of our system, we conducted experiments focusing on manual dishwashing. With a single human demonstration recorded in a mockup kitchen, the system achieved 50-100\% success for each step and up to a 40\% success rate for the whole task with different objects in a home kitchen. Videos are available at \url{https://robot-dishwashing.github.io}.
\end{abstract}


\section{Introduction}

A prominent trend in current robot learning research emphasizes the collection of extensive and varied datasets~\cite{padalkar2023open, walke2023bridgedata}, with the underlying premise that with a sufficiently large and diverse training set, an agent can autonomously extrapolate actions from data. Achievements in computer vision~\cite{radford2021learning, SAM} and natural language processing (NLP)~\cite{brown2020language, zhang2023gpt, touvron2023llama} underscore the effectiveness of this approach. However, the application of these insights in robotics presents distinct challenges. Unlike vision or NLP, where sizable datasets can be readily sourced from the web, obtaining comparable volumes of diverse data for robot interactions is exceedingly difficult, if not impossible. 

The inefficiency of robot learning methods in handling data stems from their failure to incorporate prior knowledge, relying solely on neural networks to infer everything from the available data. In contrast, when a human observes a demonstration, we typically possess a clear understanding of the task's abstractions. For example, washing a bowl involves picking it up, rinsing it, and then placing it down. However, robot cannot directly execute these human actions. Can we enable robots to learn in a manner similar to humans, \textit{i.e.} learn efficiently from demonstrations, or ideally, to acquire proficiency in a new task from a single demonstration?


The main challenge is how to interpret a demonstration into generalizable and executable abstractions for robots. Fundamental studies in robot manipulation have been focusing on the modeling of contact~\cite{mason1985robot, mason2001mechanics}. 
Following this principle, we segment manipulation tasks based on hand-object contact states: making contact, maintaining contact, and breaking contact. 
We develop a robot system to identify and reproduce the hand-object and object-object contact relationships in human video demonstrations, thereby emulating certain intentions (in the form of contacts) behind human actions.

\begin{figure}[t]
	\includegraphics[width=\linewidth]
	{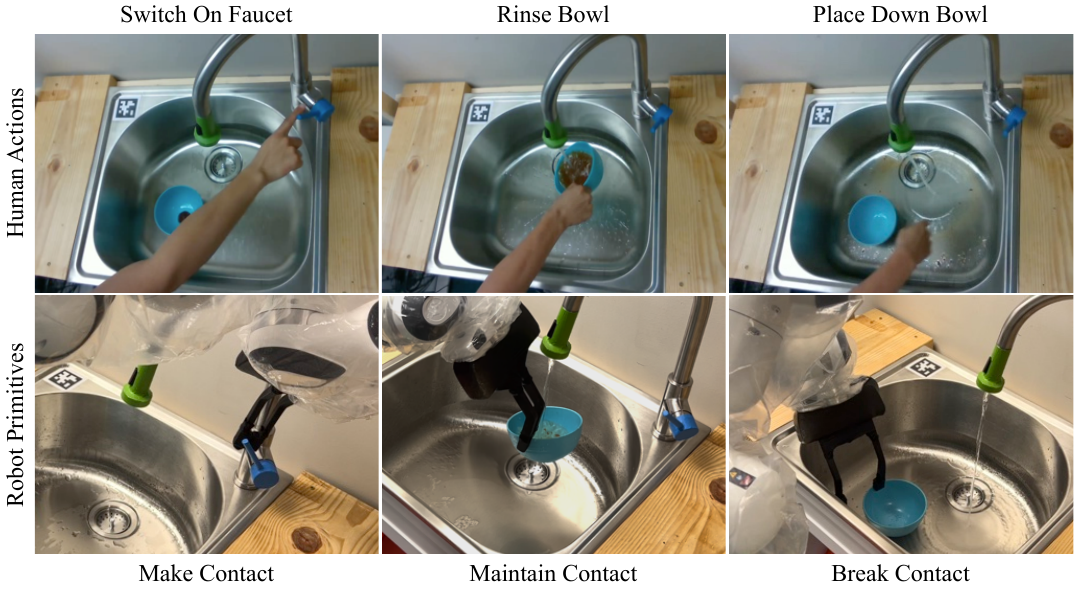}
	\caption{Human-Robot Comparative Overview of the Manual Dishwashing Task. Our system, with a single human video demonstration, learns multi-step manipulation tasks, including manipulation of articulated objects (switching on the faucet in the left column), non-prehensile manipulation (rinsing a bowl in the middle column), and pick-and-place (right column).}
	\label{fig:dishwash-teaser}
\end{figure}

We propose three high-level modules: vision, learning, and manipulation. The vision module uses multiple depth cameras to construct point clouds and estimate object poses. With this visual information, the learning module translates human actions to robot policies based on contact relationships. The manipulation module executes these policies by adapting them to current conditions, enabling a robot to replicate the observed tasks with different objects and environments.

We test our system in kitchen tasks, particularly manual dishwashing. Despite the presence of dishwashing machines, the demand for robot manual dishwashing arises from several practical considerations. First, certain items like bamboo cutting boards and delicate china are unsuitable for machine washing. Furthermore, the prevalent practice of pre-rinsing dishes before placing them in a dishwasher highlights the relevance of robotic assistance in these tasks. Additionally, dishwashing serves as a challenging multi-step manipulation task for research, encompassing three kinds of fundamental manipulation tasks (see Fig.\,\ref{fig:dishwash-teaser}): manipulation of articulated objects, non-prehensile manipulation, and pick-and-place. It also involves the manipulation of liquids and different kinds of objects.

Our system shows the generalizability in experiments. We provided demonstrations in a mockup kitchen in our lab and tested the system under varied conditions in a home kitchen, differing in geometry and appearances from the learning environment. The system achieved 50-100\% success for each step of the dishwashing task and up to a 40\% success rate for the whole task with different objects and environments.

The main contributions are:
\begin{enumerate}
\item A system that can learn and execute multi-step manipulation tasks from a single human demonstration. This involves processing RGBD videos to identify task-relevant key poses of objects, leveraging the generalizability of a large vision model.
\item A data efficient solution for robots to replicate human actions, considering the human-robot differences in kinematics and collision geometry.
\item Evaluations of the robot system in real-world experiments, providing practical evidence of its effectiveness and applicability.
\end{enumerate}


\section{Related Work}

\subsection{Learning From Demonstration}
Extensive work in the field of Learning from Demonstration (LfD) typically involves human supervision through methods like teleoperation (using a joystick or VR interface)~\cite{zhang2018deep, zhao2023learning, arunachalam2023holo} or kinesthetic teaching~\cite{calinon2007learning, levine2016end}, where a user physically guides the robot arm. However, collecting demonstrations with these approaches can be laborious and time-consuming.

Recent developments have explored alternative methods for providing human demonstrations, such as retargeting hand-pose estimation to a robot end effector~\cite{arunachalam2023dexterous, qin2022dexmv, telekinesis} and training policies directly from first and third-person human demonstrations~\cite{song2020grasping, wang2023manipulate, BharadhwajVisual}. These approaches aim to replicate human actions by robots, ignoring human-robot differences.
The approaches most similar to ours emphasize objects in demonstrations \cite{wen2022you, zhu2023learning, zhu2022viola}, but these approaches do not apply to complex multi-step tasks that involve contact-rich interactions over extended periods. Addressing the aforementioned limitations, our system centers on the contact dynamics between the human hand and objects, as well as between objects themselves. Our work enhances generalizability and data efficiency, empowering robots to learn multi-step manipulation tasks effectively from a single demonstration.

\subsection{Object Pose Estimation}
We rely on the recovery of object poses to provide robots with an understanding of the scene. We broadly categorized methodologies into four types: Point Pair Features based methods~\cite{vidal2018method, drost2010model, pan2023tax}, Template Matching~\cite{hodan2017t, devgon2020orienting}, Learning-based approaches~\cite{wang2021gdr, hu2022perspective, wu2022keypoint, su2022zebrapose}, and methods utilizing 3D Local Features~\cite{hagelskjaer2020pointvotenet}. Other works seek to enhance pose estimation accuracy using multiple views~\cite{chao2021dexycb, labbe2020cosypose, shivakumar2019ho}. However, these all require additional training for adaptation to unseen data and suffer from limited generalization to objects and scenarios deviating from their training data.

Recent progress includes using Transformers~\cite{amini2022yolopose, zhang2022trans6d, huang2020hot, zou20226d} and Neural Radiance Fields (NeRF)~\cite{yen2021inerf, linerf}. Ongoing developments are regularly reported in the Benchmark for 6D Object Pose Estimation~\cite{sundermeyer2023bop}. Yet, applying these methods to customized environments remains challenging because they are specifically designed for standard datasets and benchmarks, and are not tested for generalization to data that is similar to, but different from, the training environment.

Addressing these limitations, we employ Grounded Segment Anything~\cite{GSAM} for object detection and segmentation from images with text prompts, coupled with Iterative Closest Point (ICP)~\cite{ICP} to determine object pose. This approach offers generalizability, functioning effectively in diverse scenarios without additional training. Moreover, this part of our system is modular, allowing for replacement as more advanced techniques emerge, ensuring adaptability and future-proofing.

\subsection{Temporal Video Segmentation}
Temporal video segmentation is essential in our work for segmenting videos into executable robot primitives. There are many datasets in this domain, some incorporating spatio-temporal annotations and object relations~\cite{ji2020action}, which mainly focus on bounding boxes. Seminal video datasets~\cite{perazzi2016benchmark, wang2021unidentified, YouTubeVOS} offer pixel labels over time, yet these are typically short-term and lack fine-grained action labels. The Epic-Kitchen dataset~\cite{darkhalil2022epic}, collected using an ego-centric camera, is particularly relevant to our research. It has been the basis for many video segmentation tasks, including Hand Object Segmentation~\cite{VISOR2022, shan2020understanding}, which aims to identify contact relationships between hands and objects in images. This dataset also supports experiments in tasks including action recognition~\cite{caelles2017one, perazzi2017learning, yang2018efficient, voigtlaender2017online} and action detection~\cite{feichtenhofer2019slowfast, lin2019bmn}. However, these methods generally do not surpass 60\% in accuracy.

To tackle this challenge, we opted for an approach using RGBD videos: by leveraging accurate image segmentation, we reconstruct object point clouds and calculate contact relationships to segment robot primitives. This approach eliminates the dependency on a particular dataset and provides accurate segmentation points for robots to understand and replicate human actions.

\section{Problem Formulation}

The problem is we need a robot system to learn and execute corresponding robot primitives with human actions observed from a single demonstration. These primitives need to be adaptable to various objects and environments, differing from the learning scenario, and executable by different robot arms when provided with task-specific parameters. Therefore, the goal is to develop a function \( g \) that maps each frame \( f_i \) in a video \( V \) to a robot primitive \( p \) in the set of primitives \( P \). Formally,
\begin{equation*}
	g: F \times O \rightarrow P\text{,}
\end{equation*}
where \( F \) represents the set of frames in the video, and \( O \) denotes the set of task-relevant objects. 

The set \( P \) could be pre-defined and needs to span the range of \( A \), all possible actions observable in human demonstrations. A fundamental feature of these primitives is their clear demarcation, designed to avoid overlap, as the robot only executes one primitive at one time. In this case, each action \( a \) maps to a unique primitive \( p \) through a function \( \phi \), leading to the following definition of \( P \):
\begin{equation*}
	P = \{ p \mid p = \phi(a), \forall a \in A \}\text{.}
\end{equation*}

We make three assumptions. (1) All task-relevant objects and environments are visible in at least one frame of the video. (2) The objects can be treated as rigid, even if not rigid in nature. (3) The scenario involves only one instance of each object type, eliminating the complexity of object tracking.


\section{Methods}
\begin{figure*}[t]
	\includegraphics[width=\linewidth]
	{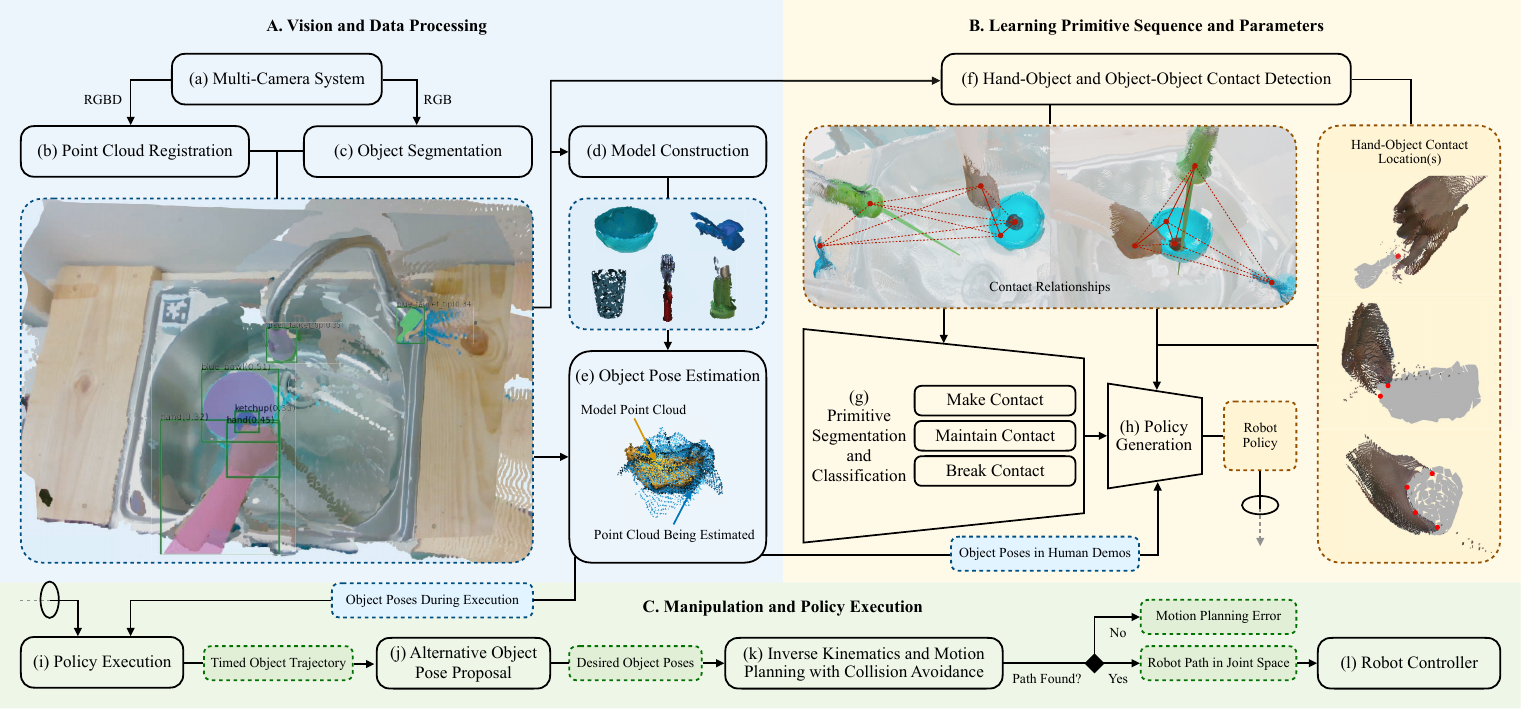}
	\caption{
	System Overview. The \textbf{vision} module processes RGBD images captured from \textit{(a)~Multi-Camera System} into a \textit{(b)~Point Cloud Registration}. With \textit{(c)~Object Segmentation}, we convert point clouds to models (\textit{(d)~Model Construction}) for \textit{(e)~Object Pose Estimation}. In the \textbf{learning} module, we calculate distances between object point clouds for \textit{(f) Hand-Object and Object-Object Contact Detection}, and we use changes in contact relationships to guide \textit{(g) Primitive Segmentation and Classification} and \textit{(h)~Policy Generation}. The \textbf{manipulation} module executes robot policy (\textit{(i) Policy Execution}) to generate a timed object trajectory. With adjusted desired object poses from  \textit{(j) Alternative Object Pose Proposal}, \textit{(k) Inverse Kinematics and Motion Planning with Collision Avoidance} finds a robot path and sends joint angle commands to the \textit{(l) Robot Controller}.}
	\label{fig:dishwash-system-1}
\end{figure*}

The system (Fig.~\ref{fig:dishwash-system-1}) has vision (Sec.~\ref{secVision}), learning (Sec.~\ref{secLearn}), and manipulation (Sec.~\ref{secManipulation}) modules to achieve its adaptive capabilities. The vision module processes RGBD videos from a human demonstration, forming point clouds and detecting object poses. The learning module then uses this information, coupled with predefined primitives, to understand and generalize the intent behind the demonstrated actions. The manipulation module executes robot policy and controls the robot, allowing the system to replicate the observed task in varied environments and with different objects.

\subsection{Vision Module}
\label{secVision}
The vision module serves as a bridge between raw visual input and actionable data for other modules of the system. It processes RGBD images captured from \textit{(a)~Multi-Camera System} into a \textit{(b)~Point Cloud Registration}. With \textit{(c)~Object Segmentation}, the vision module tags each point of the point cloud to a corresponding object or environment, and subsequently transforms the points into meshes, as described in \textit{(d)~Model Construction}. These point clouds and meshes serve as models in \textit{(e)~Object Pose Estimation} and collision detection in the manipulation module.

\textbf{\textit{(a) Multi-Camera System:}}
We use eight RealSense D435 depth cameras for their convenience and compatibility with our requirements. Calibration is the first step to enable accurate depth perception and spatial understanding. We use Multical~\cite{MultiCal}, paired with a customized April Tag~\cite{AprilTag} board to perform the calibration process. Throughout the human demonstration phase, all eight cameras operate in synchrony, capturing $640 \times 480$ at 30\,fps videos.

\textbf{\textit{(b) Point Cloud Registration:}}
We use the color and depth information from the recordings, coupled with calibrated camera intrinsics and extrinsics to construct point clouds. These point clouds amalgamate data from all eight cameras to provide a comprehensive and cohesive spatial representation. To establish a consistent reference coordinate frame, we set the origin of the point cloud at a specific fixed camera, ensuring standardization and ease of data interpretation while the system learns and executes tasks.

\textbf{\textit{(c) Object Segmentation:}}
We conduct image segmentation on captured color images, providing a mask for each object of interest. To achieve this segmentation, we use an off-the-shelf method called Grounded Segment Anything~\cite{GSAM}, a combination of Grounding DINO~\cite{liu2023grounding} and Segment Anything~\cite{SAM}. This method detects objects using text prompts and produces detailed object masks. Notably, during the recording of human demonstrations, we define the objects of interest through text prompts for detection.

\textbf{\textit{(d) Model Construction:}}
To capture the environment's point cloud, we focus on the initial frames of the demonstration when the scene is devoid of any objects. We refine the point clouds of both the environment and objects by removing outliners that are further away from their neighbors on average. These refined point clouds serve as models for pose estimation. 

Next, we transform the point clouds into meshes through the Poisson surface reconstruction method~\cite{kazhdan2006poisson}. This method retains points in point clouds as the vertices of a resulting triangle mesh. We then decompose the reconstructed surfaces into convex components, using the V-HACD technique~\cite{VHACD}. These meshes, less noisy and more computationally efficient than original point clouds, serve as models for collision detection, enabling the safe and efficient operation of the system in real-world scenarios.

\textbf{\textit{(e) Object Pose Estimation:}}
Estimating the pose of an object is a part of understanding its spatial context. For each frame in recorded video, we compute relative poses between a segmented point cloud and the model point cloud of the object, using Iterative Closest Point (ICP)~\cite{ICP}. ICP, as it iteratively refines the alignment of two point clouds, is effective in this context, where capture point clouds represent similar objects under similar conditions. By minimizing the distance between corresponding points in these point clouds, ICP can provide accurate object pose estimation.

During robot execution, the vision module detects the presence of objects and discerns their poses. When there is a need to determine an object's pose, all cameras synchronously capture RGBD images, and we determine the pose using the same procedure as in processing demonstration videos. The manipulation module uses this pose as a bridge to align the execution reality with human demonstrations.

\subsection{Learning Module}
\label{secLearn}
The primary goal of the learning module is to convert human demonstrations into robot primitives, emphasizing the understanding and recovery of contacts in manipulation tasks. In this process, we calculate distances between object point clouds for \textit{(f) Hand-Object and Object-Object Contact Detection}, and we use changes in contact relationships to guide \textit{(g) Primitive Segmentation and Classification}. By combining object poses from human demonstrations with contact information, \textit{(h)~Policy Generation} formulates a robot policy for execution in the manipulation module.

\textbf{\textit{(f) Hand-Object and Object-Object Contact Detection:}}
We assess the distances between two point clouds to determine their contact relationship. To minimize noisy fluctuations in contact change, especially when the distance is close to the threshold, we use two distinct thresholds in a sequential test. The threshold for breaking contact is higher than that for making contact. Additionally, we conduct this test in reverse order. If the contact results vary when evaluated forward and in reverse, we choose the outcome that leads to the fewest contact changes.

Moreover, we record the contact location(s) between the hand and the object it holds, which could be useful in generating a robot policy for making contact with the object. In the reference frame of the object's model point cloud, we cluster points that are close to the hand, and the center of each cluster represents a contact location.

\textbf{\textit{(g) Primitive Segmentation and Classification:}}
We compartmentalized human actions into three generalizable robot primitives, each rooted in the nature of hand-object contact relationships:

\begin{itemize}
\item \textbf{Make Contact}: This represents the initial point of interaction between the hand and the object. For instance, it correlates to the human action of picking up an object.
\item \textbf{Maintain Contact}: This primitive encompasses actions where the hand maintains contact with an object over a prolonged period, manipulating it in various ways. An illustrative example would be rinsing a bowl while continuously holding it.
\item \textbf{Break Contact}: This denotes the cessation of hand-object interaction. Typically, this corresponds to placing or letting go of an object.
\end{itemize}

We posit that we can distill most human manipulation actions into these three categories, under the assumption that all objects are rigid bodies. Fig.~\ref{fig:dishwash-traj} presents an example of breaking down the task of washing a bowl. 

The boundaries between primitives can sometimes be blurred. For example, the action of switching a faucet on or off may involve making and breaking contact with the faucet within a few frames. However, we generate robot policy relying on contact changes, not on these blurred boundaries. 

\textbf{\textit{(h) Policy Generation:}}
The robot policy for a task consists of a sequence of robot primitives, mirroring those in human demonstrations. Each primitive requires learned parameters representing the information learned from the demonstration, as well as execution parameters provided when the manipulation module executes the primitive.

Making contact requires hand-object contact location(s) as the learned parameters, with the goal of having a robot make contact with an object similarly to human demonstrations. If a human makes contact with an object at more locations than the two fingers of a robot can accommodate, as illustrated by holding a cup in Fig.~\ref{fig:dishwash-system-1}\,(f), we manually define applicable contact locations for the object.

The learned parameter of the `breaking contact' primitive is the object's final pose at the moment it breaks contact with the hand. We can override this with a manually defined object pose in cases where we do not want the robot to place the object at a location similar to that in the human demonstration.

While the hand maintains contact with an object, changes in object-object contact relationships and the object's appearance and disappearance are crucial. The learned parameters include object poses and contact relationships captured in the demonstration whenever these changes occur. The aim is to have the robot replicate the relative object pose with the same contact relationships.

\subsection{Manipulation Module}
\label{secManipulation}
As described in \textit{(i) Policy Execution}, the manipulation module executes robot policy by providing execution parameters to primitives and generates a timed object trajectory. However, it is challenging for a robot to move an object precisely along this trajectory because of its different kinematics and collision geometry compared to a human. To tackle this issue, we have implemented \textit{(j) Alternative Object Pose Proposal} function, which suggests adjustments to desired object poses, thus offering more possibilities for \textit{(k) Inverse Kinematics and Motion Planning with Collision Avoidance} to find a path for the robot in joint space. The \textit{(l) Robot Controller} then takes these joint angle commands to control the robot's movements.

\textbf{\textit{(i) Policy Execution:}}
The robot policy consists of a sequence of robot primitives that take learned parameters from demonstrations and execution parameters in the testing environment. Provided with current object and environment states, robot primitives generate a ``timed object trajectory'' for execution. This trajectory, comprising a series of timestamped object poses relative to each other, translates the contact relationships and temporal evolution of the object's poses, as learned in the demonstration, into the execution scene. This allows for the generalization and replication of human-demonstrated actions, irrespective of the specific robots or environmental conditions.

\begin{figure*}[t]
	\includegraphics[width=\linewidth]{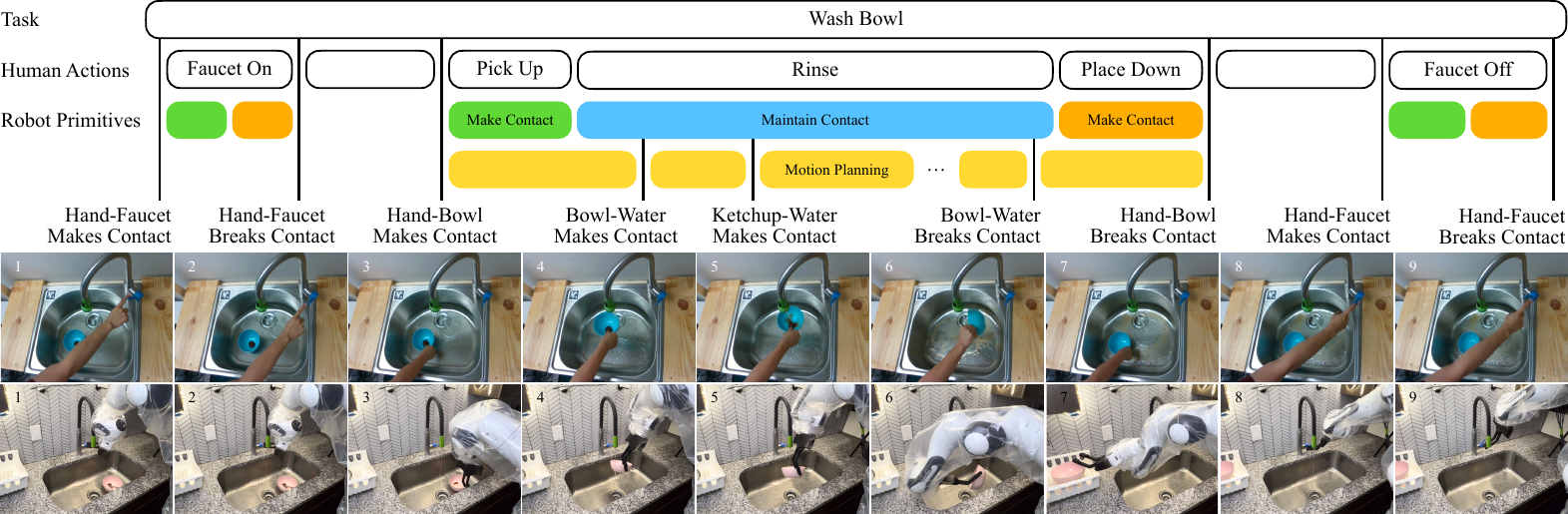}
	\caption{Overview of the Task of Washing A Bowl. The system segments human actions into robot primitives (top). We recorded demonstrations in the lab kitchen (middle row) and tested the system in the home kitchen (bottom row).}
	\label{fig:dishwash-traj}
\end{figure*}

\textbf{\textit{(j) Alternative Object Pose Proposal:}}
As humans and robots are different in their anatomical and mechanical configurations, a robot cannot directly copy human demonstration motions. For example, a human can flip the wrist to pour water out of a bowl, but since a robot's wrist is much larger, it needs to lean over with the whole arm to flip the bowl, carefully planning its motion to avoid colliding with the sink (see the 5th and 6th images in the middle and bottom row of Fig.~\ref{fig:dishwash-traj}). Sometimes it is even impossible for the robot to find any motion to replicate a demonstrated object pose. Therefore, we have implemented a function that proposes alternate desired object poses, increasing the possibility for the robot to find a feasible motion solution. This function rotates an object a full circle around its symmetric axis (if present) and small angles around its non-symmetric axes, while maintaining the contact relationships among interacting objects.

\begin{figure}[t]
	\includegraphics[width=\linewidth]{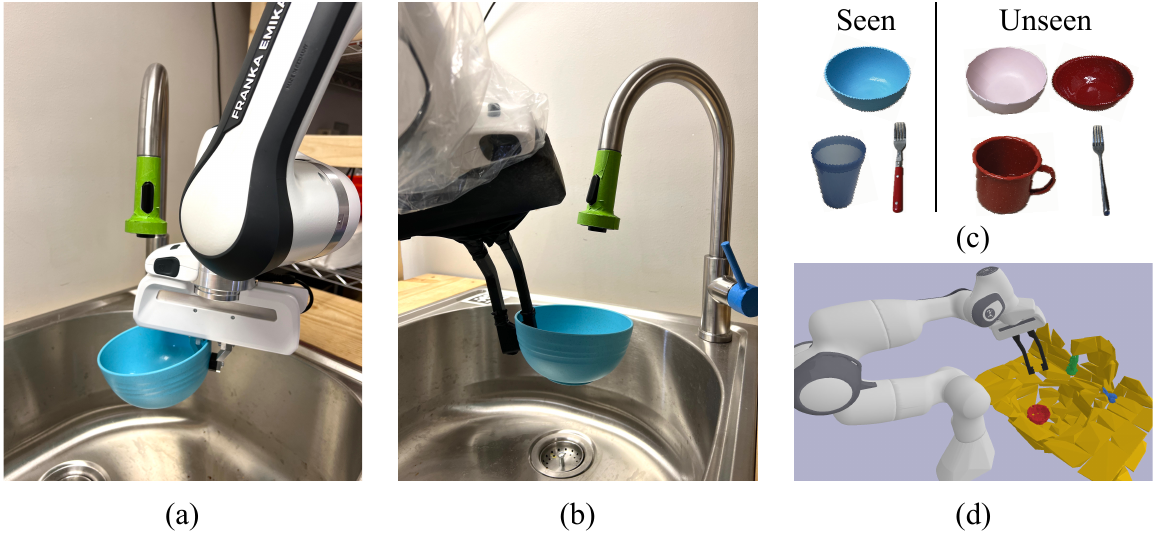}
	\caption{Robot, Objects, and Simulation Environment. The (a) original Franka Hand fingers are so short that the robot blocks water. We solved this by designing (b) longer fingers. We waterproof the robot with poly tubing on its arm and a glove on the gripper. We use objects in (c) for testing. We use (d) a simulated environment in PyBullet for collision checking and motion planning.}
	\label{fig:dishwash-robot}
\end{figure}

\textbf{\textit{(k) Inverse Kinematics and Motion Planning with Collision Avoidance:}}
The robot needs to find a collision-free path to reach one of the desired object poses suggested by the \textit{(j) Alternative Object Pose Proposal} function. We import object and environment meshes constructed by the vision module into Pybullet~\cite{PyBullet} for collision checking (see Fig.~\ref{fig:dishwash-robot}\,(d)). We use RRT-Connect~\cite{rrtconnect} to compute the path between IK poses solved by TRAC-IK~\cite{TRACIK}. If there is no path found after looping through all possible desired object poses, the system exits with a motion planning failure.

\textbf{\textit{(l) Robot Controller:}}
We match the robot's path duration with the timed object trajectory. We apply the method from~\cite{kunz2012time} to obtain a robot path that minimizes the time and is feasible for the robot to follow under robot velocity and acceleration limits. If this optimized path's duration exceeds that of the object trajectory, we adopt it as is. Otherwise, we extend the duration of the robot path to align with the time specified in the object trajectory.

We employ the Proportional-Derivative (PD) joint position controller, as implemented in Polymetis~\cite{Polymetis2021}, for robot control. The robot path is upsampled to joint angle commands at 1000~Hz and then transmitted to the robot through a computer equipped with a real-time kernel.

\section{Experiments and Results}
We present a series of experiments to test the robot system's effectiveness in different learning and testing environments (Sec.~\ref{sec:Env}). We waterproofed a Franka Research 3 robot and customized its gripper to adapt to the dishwashing task (Sec.~\ref{sec:Robot}). Sec.~\ref{sec:Results} are the results.

\subsection{Learning and Testing Environments}
\label{sec:Env}
We recorded human demonstrations in a mockup kitchen in our lab (see the middle row of Fig.~\ref{fig:dishwash-traj}). This setting facilitated safe testing, especially in managing water usage during robot experiments. After testing the system in the same environment as the demonstrations, we moved it to a home kitchen (see the bottom row of Fig.~\ref{fig:dishwash-traj}), providing more challenging and varied settings to test its adaptability in the real world. 

We covered the faucet lever with blue tape and the water outlet with green tape to help the vision module distinguish these two parts. Since Grounded Segment Anything does not segment water and we treat water flow as a rigid body, we deduce the water flow as a cylinder at the water outlet location.

\subsection{Robot Setup}
\label{sec:Robot}
We used a Franka Research 3 robot arm equipped with a Franka Hand for experiments, enhancing it with customized 3D-printed fingers. The original fingers of the Franka Hand were too short, causing the robot's hand and wrist to block water flow under a faucet (see Fig.~\ref{fig:dishwash-robot}\,(a)). Therefore, we designed longer fingers with a 135-degree angle (see Fig.~\ref{fig:dishwash-robot}\,(b)), positioning the robot's wrist away from the manipulated object and allowing unimpeded water flow during dishwashing.

In the experiments, the presence of water necessitated waterproofing the robot. As shown in Fig.~\ref{fig:dishwash-robot}\,(b), we encased the robot arm in poly tubing and sheathed the gripper in a latex glove. This design is practical and cost-effective; the chosen materials not only maintain the robot's operational flexibility and dexterity but also offer ease of replacement and maintenance, thereby ensuring the robot's longevity in wet environments.

\subsection{Results}
\label{sec:Results}
We designed experiments to test the adaptability of our system across a variety of scenarios, including interactions with both seen and unseen objects, their initial locations, and different environments. The results in Table~\ref{table:dishwash} illustrate the system's performance. The seen environment was the lab kitchen, used for human demonstrations, and the unseen environment was the home kitchen, which inherently implied different initial locations for all objects. We tested with bowls, cups, and forks (see Fig.\,\ref{fig:dishwash-robot}\,(c)). We demonstrated washing ketchup off a blue bowl and the system generalized this to a pink bowl with ketchup and a red bowl with mustard. It also adapted to different cups and forks and to different initial rotation angles of faucet levelers in two environments.

\begin{table}[]
\caption{Success rate of dishwashing task tested on \textbf{S}een and \textbf{U}nseen \textbf{Obj}ects, \textbf{Loc}ations, and \textbf{Env}ironments.}
\centering
\begin{threeparttable}
\begin{tabular}{ccc>{\centering\arraybackslash}p{7mm} ccccc}
\toprule
Obj & Loc & Env & F-On\tnote{1} & F-Off\tnote{1} & Pick & Place & Rinse & All\tnote{2} \\ 
\cmidrule(lr){1-3} \cmidrule(lr){4-8} \cmidrule(lr){9-9}

S  & S    & S & 0.9  & 0.8    & 0.8 & 1.0 & 0.8 & 0.4\\
U  & S    & S & --   & -- & 0.8 & 1.0 & 0.4 & 0.2\\
S  & U    & S & 0.6  & 0.5    & 0.6 & 0.8 & 0.8 & 0.2\\
U  & U    & S & --   & -- & 0.4 & 0.8 & 0.6 & 0.2\\
S  & U    & U & --   & -- & 0.8 & 0.8 & 0.8 & 0.4\\
U  & U    & U & 0.5  & 0.5    & 0.6 & 1.0 & 0.6 & 0.2\\
\bottomrule     
\end{tabular}

\begin{tablenotes}
\item[1] F-On: Switch On Faucet. F-Off: Switch Off Faucet. We only tested with one faucet in each environment. \textsuperscript{2}\,All five steps at once.
\end{tablenotes}

\end{threeparttable}
\label{table:dishwash}
\end{table}

In the lab kitchen, with objects positioned as in the human demonstrations, the system achieved a success rate of at least 80\,\% for each step over five trials. However, we observed that errors in individual step accumulated over time, leading to an overall success rate dropping to 20-40\,\% when the robot attempted to complete the entire task.

When we modified the experiment conditions by moving objects to different locations, the system's performance further declined, with the success rate for each step falling to 40-80\,\%. This reduction in success rate affected the robot's ability to complete the full tasks, with a success rate of only 20\,\%.

In the unseen environment (home kitchen), additional challenges arose for the system, particularly for the vision module. The system had difficulties accurately determining the pose of the faucet lever, resulting in a decreased success rate for switching the faucet on and off to 50\,\%. Despite achieving similar success rates in picking, placing, and rinsing steps as in the lab kitchen, the robot managed to complete the entire task with a success rate of 20-40\,\%.

The analysis of failure cases in the experiments highlights two key areas for future improvement. Pose estimation inaccuracies were the most significant issue, accounting for 50\,\% of the failures. Motion planning errors, where the robot couldn't find a feasible path to complete a primitive, contributed to 26\,\% of the failures. 

We excluded trials with other types of failures from the success rate calculations, as they did not reflect our intention in designing the system. In these cases, the system had successfully replicated human motions, but the task failed due to external factors like different object materials, robot strength limits, etc. In 10\,\% of all failure cases, the robot failed to rinse off ketchup adequately. The robot dropping the bowl when it became too heavy with water accounted for 8\,\% of failures. Instances of the faucet being opened but with low water flow and other unclassified issues each caused 3\,\% of failures, with the latter including unrepeatable failures such as external disturbances during experiments. Future improvements, like enhancing the robot's mechanical design with anti-slip fingers and adding a feedback loop to recover from failures, could mitigate these issues. 

\section{Conclusions and Future Work}
This paper presents a system that learns multi-step manipulation tasks from a single human demonstration. We show that a key is the identification and replication of hand-object and object-object contact relationships. By leveraging the generalizability of a large vision model, the system translates human actions into robot primitives, and the implementation of motion planning allows a robot to replicate the demonstrated task without exactly copying human motions. Our experiments with the dishwashing task, involving various objects and environments, highlight the system's effectiveness and practicality.   

The system is a prototype with several limitations that suggest directions for future work. First, currently hindered by RealSense camera noise and calibration errors, the vision module needs upgrade to more accurate depth cameras or calibration-free 3D reconstruction algorithms like Neural Radiance Fields (NeRF) to improve the object pose estimation accuracy. Second, expanding beyond replicating rigid body poses, future development might involve simulations for soft objects to infer physics from videos. Third, the system, aiming to replicate human strategies, faces challenges in tasks like multi-fingered object handling---future work may involve a grasping planner or reinforcement learning for learning strategies or primitive sequences beyond existing demonstrations. Finally, the system generating object-centric trajectory, adaptable across different robots and environments, has only been tested for dishwashing---investigating how the idea is applicable to other tasks and robots is a promising direction for future research.



\balance
\bibliographystyle{IEEEtranS}
\bibliography{citations}

\end{document}